\documentclass{article} 
\usepackage{iclr2021_conference,times}

\usepackage{amsmath,amsfonts,bm}

\def\eqref#1{equation~\ref{#1}}

\def\1{\bm{1}}

\DeclareMathAlphabet{\mathsfit}{\encodingdefault}{\sfdefault}{m}{sl}
\SetMathAlphabet{\mathsfit}{bold}{\encodingdefault}{\sfdefault}{bx}{n}

\usepackage{hyperref}
\usepackage{url}
\usepackage{todonotes}
\usepackage{graphicx}
\graphicspath{ {./images/} }
\usepackage{subcaption}
\usepackage{float}
\usepackage{sidecap}
\usepackage{tikz}
\usepackage{amsmath}
\usepackage{booktabs}       

\title{Lightweight Long-Range \\ Generative Adversarial Networks}

\author{Bowen Li \& Thomas Lukasiewicz \\
University of Oxford\\
\texttt{\{bowen.li,thomas.lukasiewicz\}@cs.ox.ac.uk} \\
}

\iclrfinalcopy 

\begin{document}
\maketitle

\begin{abstract}
  In this paper, we introduce novel lightweight generative adversarial networks, which can effectively capture long-range dependencies in the image generation process, and produce high-quality results with a much simpler architecture. To achieve this, we first introduce a long-range module, allowing the network to dynamically adjust the number of focused sampling pixels and to also augment sampling locations. Thus, it can break the limitation of the fixed geometric structure of the convolution operator, and capture long-range dependencies in both spatial and channel-wise directions. Also, the proposed long-range module can highlight negative relations between pixels, working as a regularization to stabilize training. Furthermore, we propose a new generation strategy through which we introduce metadata into the image generation process to provide basic information about target images, which can stabilize and speed up the training process. Our novel long-range module only introduces few additional parameters and is easily inserted into existing models to capture long-range dependencies. Extensive experiments demonstrate the competitive performance of our method with a lightweight architecture.
\end{abstract}

\section{Introduction}
Generating realistic and diverse samples from high-dimensional data distributions has made much progress with the emergence of autoregressive models~\citep{oord2016pixel, van2016conditional}, variational autoencoders~\citep{kingma2013auto}, and generative adversarial networks (GANs)~\citep{goodfellow2014generative}, which greatly boost various research areas, including speech synthesis~\citep{donahue2018adversarial, binkowski2019high, engel2019gansynth}, image~\citep{karras2017progressive, zhang2019self, karras2019style} and video generation~\citep{clark2019adversarial, mahon2020knowledge}, text-to-image generation~\citep{zhang2018stackgan++, xu2018attngan, li2019controllable}, text-guided image manipulation~\citep{dong2017semantic, nam2018text, li2020manigan, li2020lightweight}, and image-to-image translation~\citep{zhu2017unpaired, park2019semantic, li2020image}. 

In fact, most of the aforementioned generative approaches depend heavily on the convolution operator to model dependencies across different regions. However, the convolution operator has a fixed geometric structure with a local receptive field, and thus long-range dependencies on non-neighboring locations can only be captured by passing through several convolution layers. Unfortunately, increasing the number of layers is undesirable for devices with limited memory storage (e.g., mobile phones), because (1) it makes the architecture more complex with numerous parameters, (2)~it can cause much trouble for optimization algorithms to effectively coordinate multiple layers to capture long-range dependencies, and (3) it may fail to keep a globally semantic consistency (e.g., structure of objects) due to its fixed geometric structure. 

\begin{figure}[t]
\centering
\begin{minipage}{0.9\textwidth}
\includegraphics[width=1\linewidth, height=0.2\linewidth]{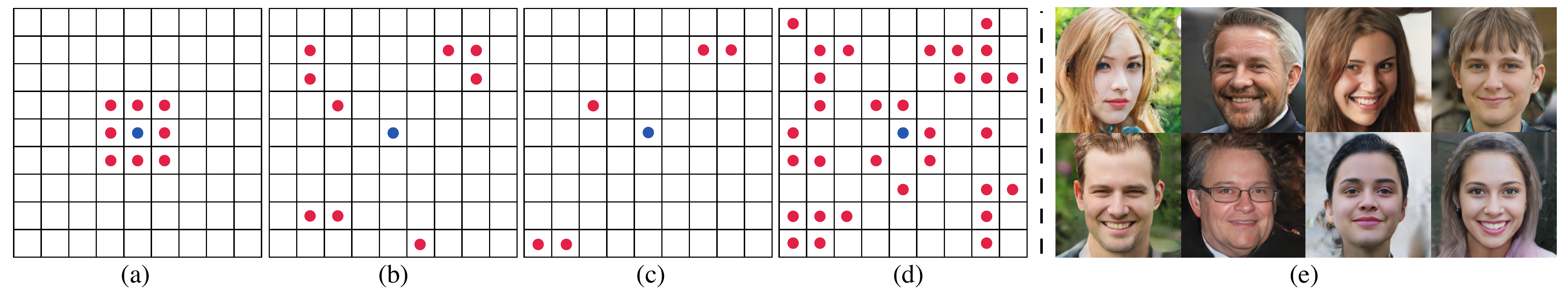}
\end{minipage}

\centering
\vspace{-2ex}
\caption{(a): local dependencies of the blue point captured by a $3\times 3$ regular convolution operator; (b): long-range dependencies of the blue point highlighted in distant locations and captured by our long-range module; 
(c) and (d) show that our long-range module can flexibly adjust the number of focused sampling locations, related to the blue point, to capture long-range dependencies; and (e)~presents eight sample results generated by our lightweight long-range network at $256 \times 256$.}
\label{fig:intro}
\vspace{-2ex}
\end{figure}

To address the above problems, it is desirable to have a module with an adjustable receptive field and augmented sampling locations to effectively capture long-range dependencies. To achieve this, we propose a novel long-range module, which enables free-form deformation of the sampling grid and allows the network to dynamically adjust the number of focused sampling pixels, shown in Fig.~\ref{fig:intro}. 
By adopting our module, the network is able to capture long-range dependencies between non-neighboring pixels without greatly increasing the network layers, enabling a possibility to install the model into memory-limited devices. 
The proposed module only introduces few additional parameters, and can be readily inserted into existing models to capture long-range dependencies.

Furthermore, we propose a novel generation strategy through which we introduce metadata into the image generation process, where the metadata can provide basic information for target images (e.g., the category and basic texture of objects), stabilize the training process, and thus free the generator from a complex architecture and speed up the training process. 

Finally, we evaluate our model on the FFHQ~\citep{karras2019style}, CUB bird~\citep{wah2011caltech}, and ImageNet~\citep{russakovsky2015imagenet} datasets, which demonstrates that our method can produce high-quality images with a great efficiency. 

\section{Related Work}

\textbf{Generative adversarial networks} have achieved much success in generating realistic images~\citep{karras2017progressive, karras2019style, karras2020analyzing}, which is widely adopted in various tasks, including text-to-image generation \citep{zhang2017stackgan, he2019attgan, li2019controllable}, image-to-image translation~\citep{chen2017photographic, isola2017image, wang2018high, li2020image}, text-guided image manipulation~\citep{dong2017semantic, nam2018text, li2020manigan, li2020lightweight}, and super resolution~\citep{sonderby2016amortised, ledig2017photo}. However, to produce high-quality results, these methods usually have a rather complex architecture with many parameters, and require a quite long time for optimization and inference, which is especially undesirable for memory-limited devices, such as mobile phones. 

\textbf{Long-range dependencies} play a critical role in deep neural networks. To capture long-range dependencies, recurrent operations~\citep{rumelhart1986learning, hochreiter1997long} are adopted for sequential data in language, and large receptive fields formed by deep stacked convolution layers are implemented for pixel data in images. Recently, self-attention~\citep{vaswani2017attention} is proposed to discard costly recurrent operations by attending to all positions in a sequence, and is widely adopted in various tasks, including machine translation~\citep{vaswani2017attention}, video classification~\citep{wang2018non}, and image generation~\citep{zhang2019self}. However, the generation of self-attention relies heavily on the implementation of the softmax function, and so almost all values in an attention map are greater than $0$. Thus, attention can only produce positive relations between pixels in an image. 
Unfortunately, not all pixels have a positive effect on others, and it is more reasonable to keep the negative effect instead of converting all relations into positive ones.

\textbf{Receptive field.} How to increase the receptive field of the convolution operator has drawn much attention.  
One complementary technique is to use dilated convolutions~\citep{chen2014semantic, yu2015multi}. With dilated convolutions, the number of parameters does not change, but the receptive field grows exponentially if the number of parameters grows linearly in successive layers. The other is to use deformable convolution~\citep{dai2017deformable, zhu2019deformable}, which learns the offset to achieve a deformation of the sampling grid. 
However, the number of sampling locations are still fixed, and the weights of deformable convolution operators are shared across all different regions. 

\section{Lightweight Long-Range Generative Adversarial Networks}
\begin{figure}[t]
\centering
\begin{minipage}{0.78\textwidth}
\includegraphics[width=1\linewidth, height=0.79\linewidth]{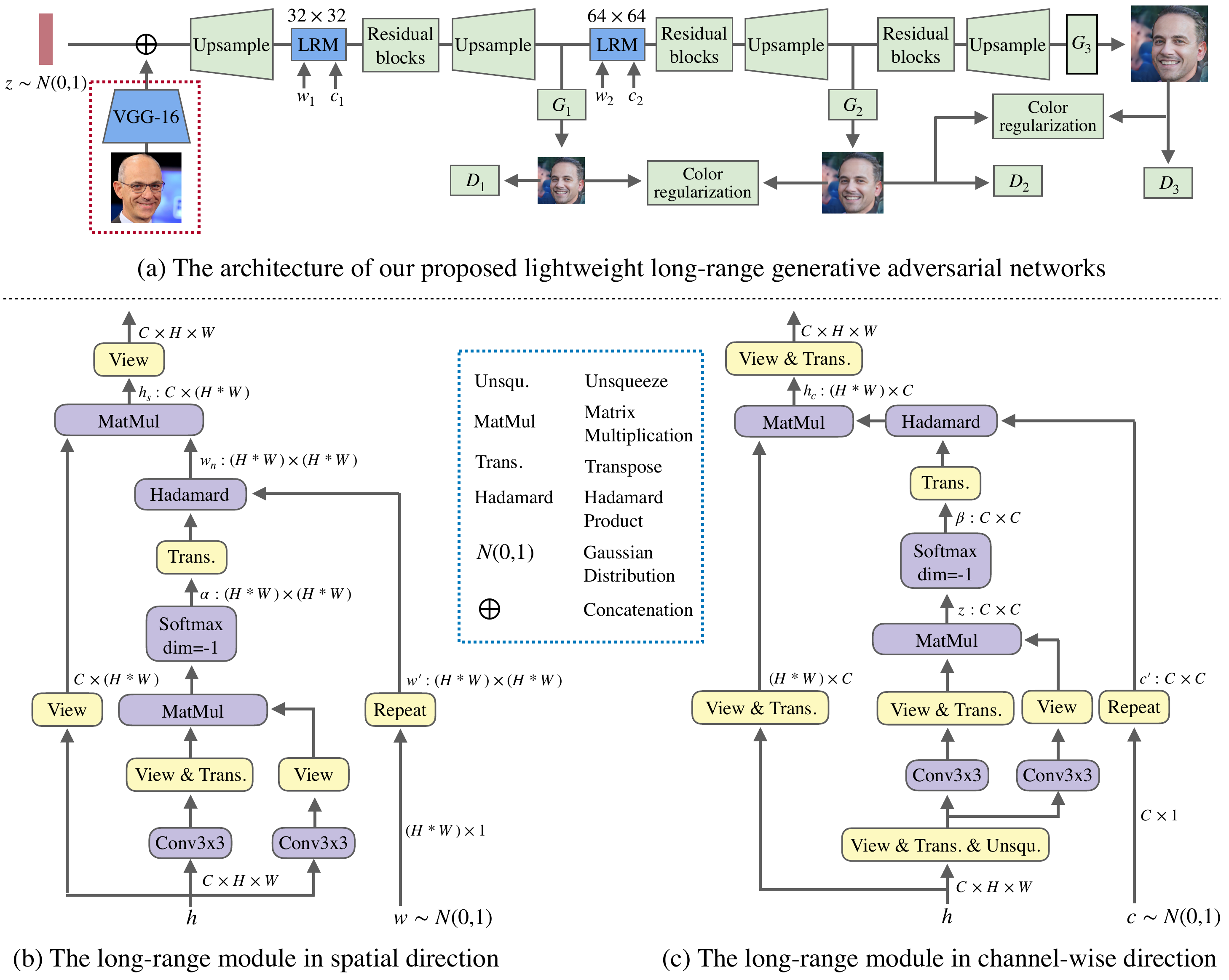}
\end{minipage}

\centering
\caption{(a): architecture of the proposed lightweight long-range generative adversarial networks; the red dashed box indicates the generation of metadata; (b): implementation of the long-range module in spatial direction; (c): implementation of the long-range module in channel-wise direction.}
\label{fig:archi}
\end{figure}

Given a random noise $z$ sampled from the Gaussian distribution $\mathcal{N} (0,1)$, our method aims to generate novel high-quality images, and at the same time, the model should be efficient and small enough for memory-limited devices. To achieve this, we propose a novel long-range module and a new generation strategy. The complete architecture is shown in Fig.~\ref{fig:archi}~(a).

\subsection{Architecture}
As shown in Fig.~\ref{fig:archi}~(a), a multi-stage architecture~\citep{zhang2018stackgan++, xu2018attngan, li2019controllable, li2020manigan} is adopted, where there are multiple generator and discriminator pairs in the model.
The reason to have such a design is based on our experimental finding that a model having a lightweight architecture with a few number of parameters is easily prone to break due to the instability of GANs. However, with the implementation of a multi-stage architecture, first, the model distribution generated from coarse low-resolution results has a better probability of intersecting with the real image distribution, enabling a better foundation for higher-quality image generation. Second, low-resolution results can also work as a regularization to constrain the randomness involved in the generation of higher-resolution images, and thus to prevent the mode collapse at higher stages. To build this constraint, color consistency regularization~\citep{zhang2018stackgan++} is adopted to keep samples generated from different generators more consistent in color. The regularization $\mathcal{L}_{C}$ at stage~$i$ is defined as:
\begin{equation}
\mathcal{L}_{C_{i}}=\frac{1}{n}\sum^{n}_{j=1}(\lambda _{1}\left \| \mu _{s_{i}^{j}}-\mu_{s_{i-1}^{j}} \right \|_{2}^{2}+\lambda _{2}\left \| \sigma_{s_{i}^{j}}-\sigma_{s_{i-1}^{j}} \right \|_{F}^{2})
\textrm{,}
\label{eq:color}
\end{equation}
where $i$ is the index of stage with $i>1$, $n$ is the batch size, $\mu_{s_{i}^{j}}$ and $\sigma_{s_{i}^{j}}$ are the mean and the covariance for the $j^{th}$ sample generated by the $i^{th}$ generator, $F$ represents the Frobenius norm, and $\lambda _{1}$ and $\lambda _{2}$ are hyperparameters. 

\subsection{Long-Range Module}
To build a lightweight network for high-quality image generation, it is desirable to have a module that can efficiently capture the long-range dependencies between non-neighboring locations rather than increasing the number of layers in a network. To achieve this, we propose a novel long-range module that can capture the long-range dependencies between distant pixels in both spatial and channel-wise directions, without introducing many additional parameters. 

\textbf{Long-range module in spatial direction.} As shown in Fig.~\ref{fig:archi}~(b), the long-range module in spatial direction takes two inputs: (1) the hidden feature $h \in \mathbb{R}^{C \times H \times W}$, where $C$, $H$, and $W$ denote the number of channels, the height and the weight of the feature map $h$, respectively, and (2) a learnable weight $w \in \mathbb{R}^{(H \ast W) \times 1}$, drawn from the Gaussian distribution $\mathcal{N}(0,1)$, where $w$ is used to perform the linear transformation on a given feature along the spatial direction.

To find the correlation between different spatial locations, we first use the convolution operator to convert the hidden features $h$ into a new semantic space to produce $h_{w1}$ and $h_{w2}$. Then, we change their shape and apply matrix multiplication between them to get a matrix $n \in \mathbb{R}^{(H \ast W) \times (H \ast W)}$, denoted as $n=h_1^{T}*h_2$, where the matrix $n$ contains all possible pairs of different spatial locations.
We further normalize it by the softmax function on the last dimension, to get a correlation value $\alpha$, 
\begin{equation}
\alpha_{i,j}=\frac{\exp(n_{i,j})}{\sum^{H \ast W}_{l=1}\exp(n_{i,l})}
\textrm{,}
\label{eq:spatial}
\end{equation}
where $\alpha_{i,j}$ denotes the extent of correlation between spatial locations $i$ and $j$, regardless of the relations being positive or negative. 
Meanwhile, we repeat the learnable weight $w$ on the column dimension to get $w' \in \mathbb{R}^{(H \ast W) \times (H \ast W)}$, where the following columns have the same values as the first one, reducing the number of learnable parameters.
Then, the spatial-relation-aware weight $w_{n}$ can be obtained by applying matrix multiplication $\times$ on $\alpha$ and $w'$, denoted $w_{n} = \alpha^{T} \times w'$, where $w_{n_{i,j}}$ represents a weight scaled by the extent of relations between spatial locations $i$ and $j$. Finally, the hidden feature $h_{s}$ capturing long-range dependencies in spatial direction can be obtained by applying matrix multiplication on $h$ with a new shape and $w_{n}$.

\textbf{Long-range module in channel-wise direction.}
As shown in Fig.~\ref{fig:archi}~(c), the long-range module in channel-wise direction takes two inputs: the hidden feature $h$, and (2) a learnable weight $c \in \mathbb{R}^{C \times 1}$, drawn from the Gaussian distribution $\mathcal{N}(0,1)$, where $c$ is used to perform the linear transformation on a given feature along the channel-wise direction.

We first use the convolution operator to convert $h$ into a new semantic space to produce $h_{c1}$ and $h_{c2}$. Then, we change their shape, followed by a matrix multiplication to get $z \in \mathbb{R}^{C \times C}$, denoted $z=h_{c1}^{T}*h_{c2}$. Finally, we normalizes $z$ on the last dimension by applying the softmax function, obtaining the correlation matrix $\beta$, 
\begin{equation}
\beta_{i,j}=\frac{\exp(z_{i,j})}{\sum^{C}_{l=1}\exp(z_{i,l})}
\textrm{,}
\label{eq:channel}
\end{equation}
where $\beta_{i,j}$ denotes the extent of positive or negative correlation between channels $i$ and $j$. We repeat $c$ on the column dimension to obtain $c' \in \mathbb{R}^{C \times C}$, and each column has the same values. Then, the channel-relation-aware weight $c_n$ is obtained by applying matrix multiplication $\times$ on $\beta$ and $c'$, denoted
$c_n = \beta^{T} \times c'$, where $c_{n_{i,j}}$ represents a weight scaled by the extent of relations between channels $i$ and $j$. Thus, the feature $h_c$ capturing the long-range dependencies in channel-wise direction is obtained by applying matrix multiplication on $h$ with a new shape and $c_n$.

\textbf{Why are the long-range dependencies important in the image generation task?}  There exist potential relations between pixels in neighboring and non-neighboring locations. These relations can work as cues to help the generator to draw images, where fine-grained details at every location are carefully coordinated with details in distant regions of the image, to keep a global semantic consistency, e.g., when a generator tries to produce an image about a cat, it is better for it to take all related pixels into account to draw different parts (e.g., eyes and mouth) at reasonable locations. 

\textbf{Why does the long-range module work better?} (1) Comparison with convolution operator: the convolution operator has a fixed geometric structure with local receptive fields, and long-range dependencies on non-neighboring locations can only be captured by passing through several convolution layers, which prevents building a lightweight model with few parameters. One possible solution is to increase the size of the convolution kernels, but it loses the computational efficiency benefited from the local convolution structure.
However, our long-range module is able to work as a complement to the convolution operator. At the cost of increasing only a few number of parameters, it helps the model to capture long-range dependencies across image regions. (2) Comparison with self-attention: self-attention is based heavily on the implementation of the softmax function, and thus almost all its values are greater than $0$. This means that self-attention utilizes the scales of positive values to highlight or ignore different image regions, i.e., giving high (or low) \textbf{positive} weights to important (or unimportant) regions. 
However, not all image regions have a positive impact on others, and some negative relations are also vital in the image generation process, especially those negative relations can work as a regularization to stabilize the training process and prevent mode collapse. To keep both negative and positive effects, our long-rang module only highlights or ignores the \textbf{relations} between different image regions, rather than the \textbf{actual values} of a hidden feature. More specifically, $\alpha$ (Eq.~\ref{eq:spatial}) and $\beta$ (Eq.~\ref{eq:channel}) can be treated as scaling weights to highlight or ignore negative and positive relations contained in two learnable $w$ and $c$, and then $w$ and $c$ are used to achieve a transformation on the hidden features.

\subsection{Generation Strategy with Metadata}
To speed up the training process, we suggest to incorporate metadata into the model to provide the generator with basic information about target images, where this information may contain cues for the desired real image distribution that the generator finally has to generate, helping it to know what kinds of objects to synthesize in advance and speeding up the training (Fig.~\ref{fig:archi}~(a), red dashed box). 

Also, to prevent the provided metadata enforcing the model to achieve the identity transformation from the meta-image, we use the global image features as the metadata, extracted from the given meta-image using a deep layer of the pretrained VGG-16 network~\citep{simonyan2014very}. Thus, the metadata only keeps summarized spatial information. The reason to have such a design is mainly because normal spatial features contain too many details about target images, such as color, shape, pose, and location of objects. Therefore, according to averaging all spatial features in each channel, those fine details can be filtered, and only elementary information is kept, ensuring a good diversity of the model.

\subsection{Objective Functions}
We train the generator and the discriminator alternatively by minimizing the generator loss $\mathcal{L}_{G}$ and the discriminator loss $\mathcal{L}_{D}$.

\textbf{Generator objective.} 
The complete generator objective has an unconditional adversarial loss and a color consistency regularization $\mathcal{L}_{C}$ (Eq.~\ref{eq:color}), where the adversarial loss encourages the generator to produce realistic fake images:
\begin{equation}
\mathcal{L}_{G}=\sum_{k=1}^{K}(-\frac{1}{2}E_{I'_{k}\sim P_{G_{k}}}\left [ \log(D_{k}(I'_{k})) \right ]) + \sum_{i=2}^{K}(\lambda_{3}\mathcal{L}_{C_{i}})
\textrm{,}
\label{eq:generator}
\end{equation}
where $K$ is the number of stages, $I'_{k}$ is the synthetic images sampled from the model distribution $P_{G_{k}}$ at stage $k$, $D_{k}$ is the discriminator at stage $k$, and $\lambda_{3}$ is a hyperparameter. 

\textbf{Discriminator objective.}
The final discriminator objective is defined as:
\begin{equation}
\mathcal{L}_{D}=\sum_{k=1}^{K}(-\frac{1}{2}E_{I_{k}\sim P_{\text{data}}}\left [ \log(D_{k}(I_{k})) \right ] -\frac{1}{2}E_{I'_{k}\sim P_{G_{k}}}\left [ \log(1-D_{k}(I'_{k})) \right ])
\textrm{,}
\label{eq:discriminator}
\end{equation}
where $I_{k}$ is the real images sampled from the true image distribution $P_{\text{data}}$ at stage $k$.

\section{Experiments}
To evaluate our method, we conduct extensive experiments on the FFHQ~\citep{karras2019style}, CUB bird~\citep{wah2011caltech}, and ImageNet~\citep{russakovsky2015imagenet} datasets, comparing with two approaches, PGGAN~\citep{karras2017progressive} and SAGAN~\citep{zhang2019self}, where PGGAN is able to produce high-quality images with a relatively simpler architecture compared with StyleGAN~\citep{karras2019style} and StyleGAN2~\citep{karras2020analyzing}, and SAGAN implements self-attention to capture long-range dependencies. Note that we do not compare our method with StyleGAN and StyleGAN2, because both approaches are based on PGGAN but have a more complex architecture with a larger number of parameters. However, the purpose of our method is to achieve a lightweight architecture with a small number of parameters, which allows the network to be implemented in memory-shortage devices.

\textbf{Implementation.} There are three stages in our model, and the scale of the output images is $256 \times 256$. However, the number of stages and the size of the output image are easily adjusted to satisfy users' preference. The hyperparameters $\lambda_{1}$, $\lambda_{2}$, and $\lambda_{3}$ are set to 1, 5, and 50, respectively. The models are trained for roughly two days on a single GPU, using the Adam optimizer~\citep{kingma2014adam} with learning rate $0.0002$.
To have the best lightweight structure and also a good performance, for the FFHQ dataset, we suggest to add our proposed long-range module to the $32 \times 32$ feature map, and for the CUB bird and ImageNet datasets, we suggest to add it to the $64 \times 64$ feature map. To have a fair comparison on the architecture and inference time, all models are implemented in PyTorch to reduce the influence from different machine learning libraries. The reproduction of these approaches is based on the source code released by authors with a careful fine-tuning. Note that to have a fair comparison, we also restrict the training time for all three methods, which is two days on a single Quadro RTX 6000 GPU.

\subsection{Quantitative and Qualitative Comparison}
\begin{table}[t]
  \centering
  \scalebox{1}{
  \begin{tabular}{l||ccc|ccc}
    \toprule
    Method & FFHQ  & CUB & ImageNet Church & IT(s) & NoP-G & NoP-D \\ 
    \midrule
    PGGAN  & 18.41  & 38.73 & 72.13 & 17.24 &  23.3M & 23.3M \\ 
    SAGAN & 79.81 & 70.93 & 84.41 & 1.89 & 13.3M & 51.6M \\ 
    \textbf{Ours} & 23.97 & 24.85 & 56.50 & 0.83 & 7.0M & 7.6M \\ 
    \midrule
    Ours w/o Meta  & 35.35 & 26.47 & 68.70 & - &  - & - \\ 
    Ours w/o LRM   & 156.59 & 42.25 & 177.28 & - &  - & - \\ 
    Ours w/ Residual   & 43.83 & 43.21 & 90.41 & - &  - & - \\ 
    Ours w/ SA   & 187.61 & 35.76 & 84.47 & - &  - & - \\ 
    \bottomrule
  \end{tabular}
  }
  \caption{Quantitative comparison: Fréchet inception distance (FID), inference time (IT) for generating 100 new results, and number of parameters in the generator (NoP-G) and the discriminator (NoP-D) of these approaches and our method on FFHQ, CUB, and ImageNet Church. ``Ours w/o Meta'' denotes without the provision of metadata. 
  ``Ours w/o LRM'' denotes without implementing the long-range module. ``Ours w/ Residual'' denotes using the residual blocks to replace our long-range module.
  ``Ours w/ SA'' denotes using self-attention instead of our long-range module. 
  For FID, lower is better. All models are benchmarked on a single Quadro RTX 6000 GPU.}
\label{table:quantitative}
\end{table}

\begin{figure}[t]
\centering
\begin{minipage}{1\textwidth}
\includegraphics[width=1\linewidth, height=0.51\linewidth]{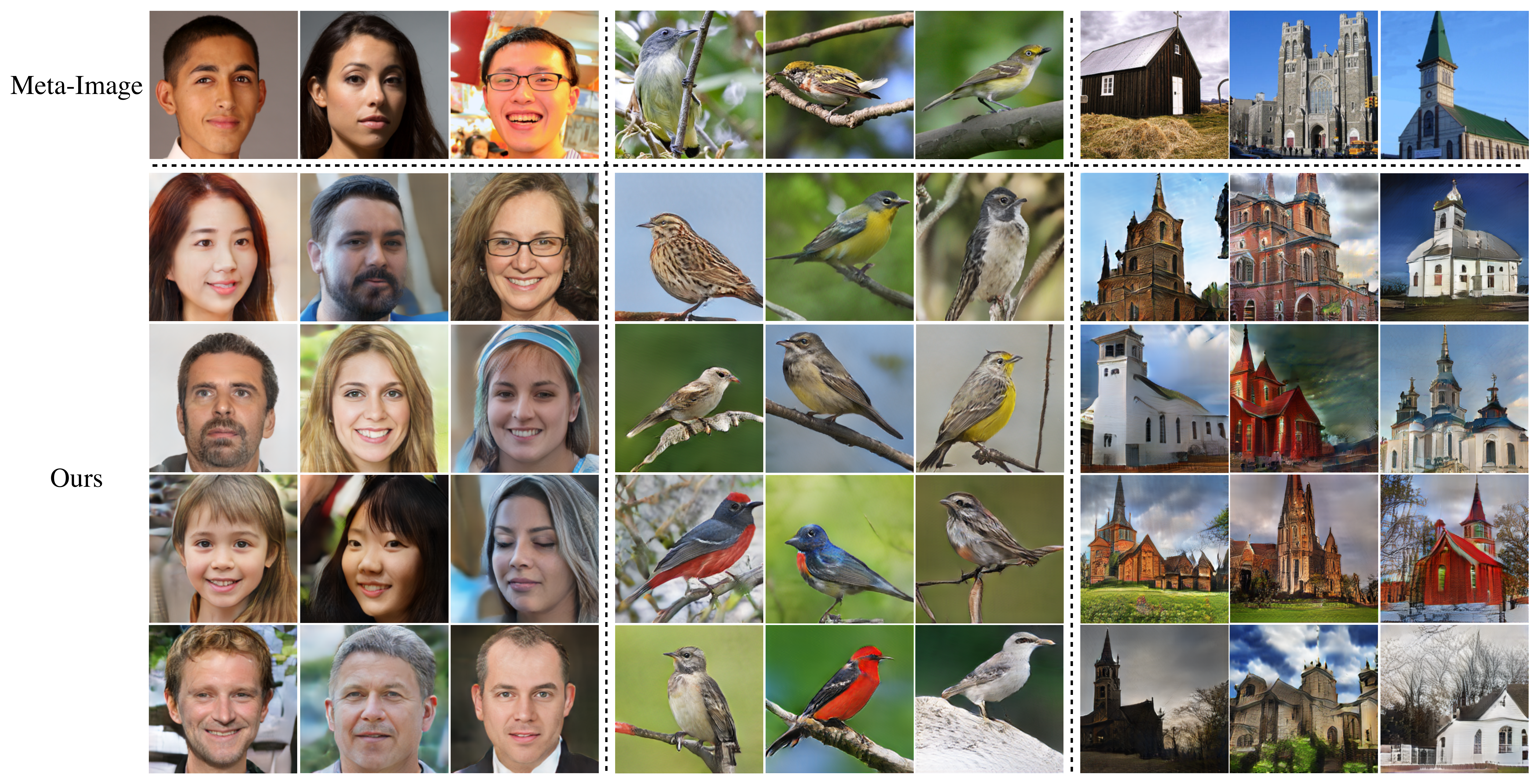}
\end{minipage}

\centering
\caption{Qualitative results generated by our method on FFHQ (left), CUB (middle), and ImageNet Church (right). The top row shows the images that are used to provide metadata for the model.}
\label{fig:qual_res}
\end{figure}

\begin{figure}[t]
\centering
\begin{minipage}{1\textwidth}
\includegraphics[width=1\linewidth, height=0.362\linewidth]{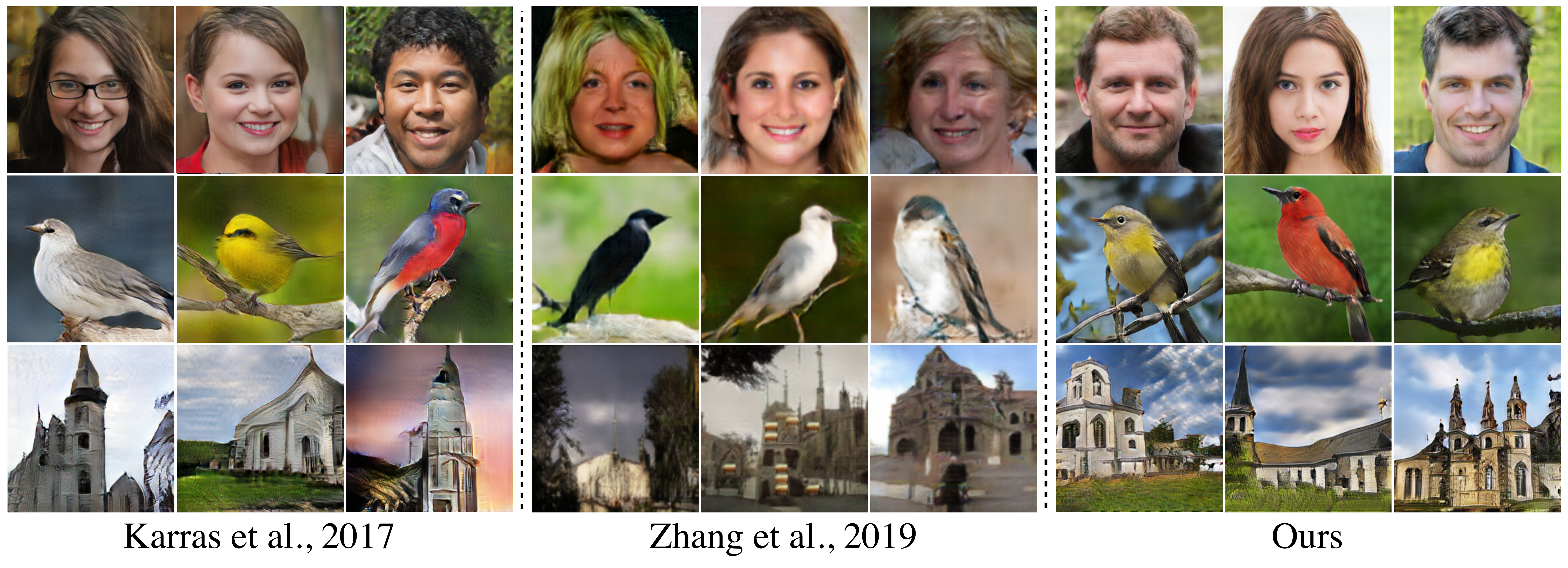}
\end{minipage}

\centering
\caption{Qualitative comparison of three methods on the FFHQ (top row), CUB (middle), and ImageNet Church (bottom) datasets.}
\label{fig:qual_cmp}
\end{figure}

\textbf{Quantitative comparison.} To evaluate the quality and diversity of synthetic images, the Fréchet inception distance (FID)~\citep{heusel2017gans} is adopted. Also, we record the inference time for generating 100 images (IT) and the number of parameters in both the generator (NoP-G) and the discriminator (NoP-D) (Million) to verify the efficiency of our method.

As shown in Table~\ref{table:quantitative}, our method has better FID values than SAGAN and competitive results compared with PGGAN. However, our method has a much smaller inference time (one order of magnitude speedup over PGGAN and two times faster than SAGAN) and fewer number of parameters. This indicates that (1) our method can generate high-quality results with a good diversity, and (2) our method is suitable for memory-limited devices, based on the size of model and inference time.

\textbf{Qualitative results.}
As shown in Fig.~\ref{fig:qual_res}, we presents example results generated by our lightweight long-range network at $256 \times 256$ along with images that are used to provide metadata for the model. As we can see, our model is able to produce high-quality images with a good diversity. Also, we can easily observe that the generated results are obviously different from the images providing metadata, and our model can even produce highly diverse results on the same meta-images, for example, the synthetic churches have a quite different structure, shape, texture, color, and background from those churches shown in meta-images. This indicates that our method can completely filter fine details contained in the meta-images, and effectively avoid copying and pasting from them.

Fig.~\ref{fig:qual_cmp} shows the visual comparison between PGGAN, SAGAN, and our method on the FFHQ, CUB bird, and ImageNet datasets. As we can see, SAGAN fails to produce realistic images at the scale $256 \times 256$ on three datasets, and compared with PGGAN, our method can achieve a competitive performance on the FFHQ dataset, but has better results on the CUB bird and ImageNet datasets. We think that the better performance on both datasets is mainly because (1) the training time is restricted, which may have a bigger impact on the performance of PGGAN and SAGAN, (2) the size of CUB and some specific categories in ImageNet are small, which may not be enough to fully optimize a heavy model with a large number of parameters. There are about 8k training images in CUB and on the average 2k images for specific categories in ImageNet tested in the paper. However, thanks to the long-range module and the provision of metadata, our model only has a small number of parameters to optimize and also gains some training cues from the metadata in advance, and (3) images in both datasets may involve the generation of multiple objects with complicated interactions rather than generally unified faces with similar texture patterns, and capturing the long-range dependencies is important in such complex image generation. This can be verified in Table~\ref{table:quantitative}: the model without the long-range module has poor FID values.  

\begin{figure}[t]
\centering
\begin{minipage}{0.86\textwidth}
\includegraphics[width=1\linewidth, height=0.44\linewidth]{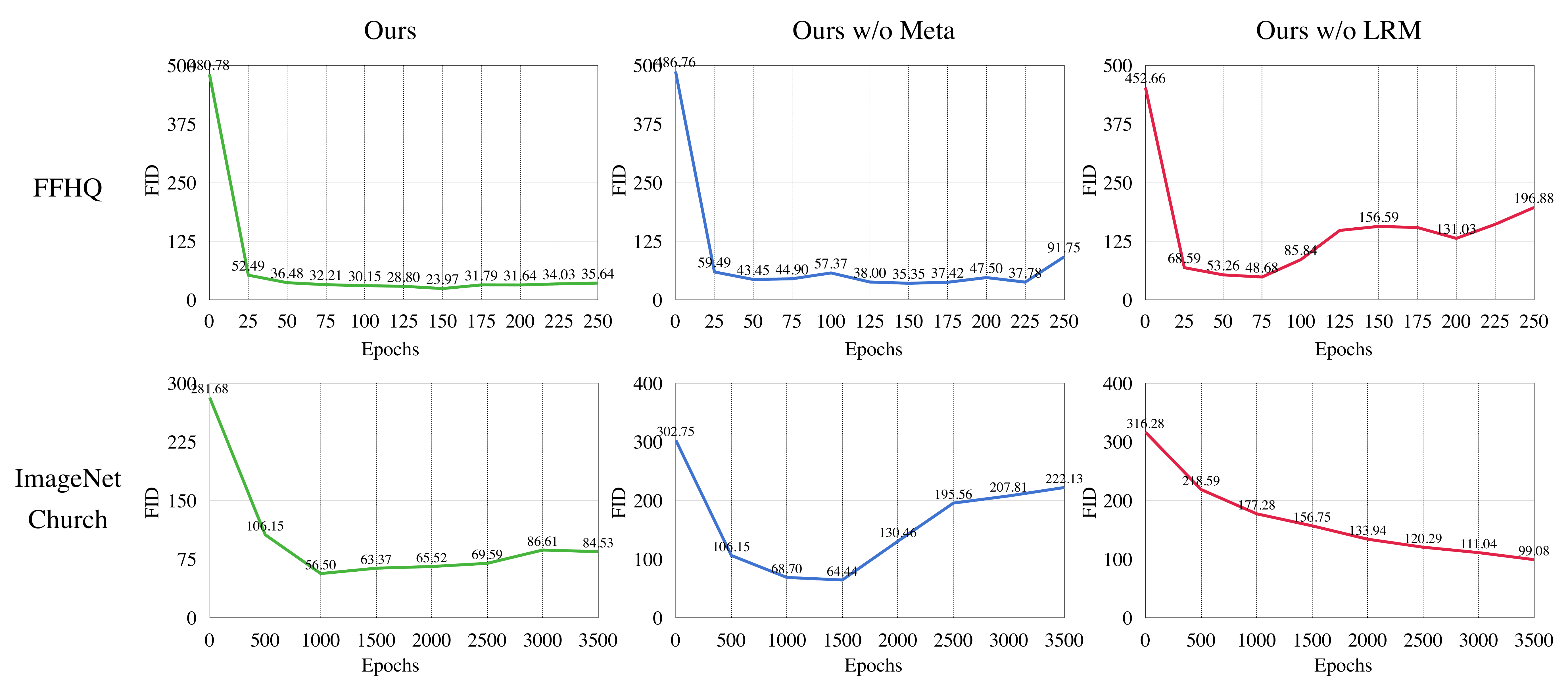}
\end{minipage}

\centering
\caption{FID values at different epochs on the FFHQ (top) and ImageNet Church (bottom) datasets.}
\label{fig:plot}
\end{figure}

\subsection{Ablation studies}
\textbf{Effectiveness of the metadata.}
To verify the effectiveness of the metadata, we first conduct an ablation study, shown in Table~\ref{table:quantitative} and Fig.~\ref{fig:plot}. As we can see, with the implementation of metadata, the FID values are improved. Also, shown in Fig.~\ref{fig:plot}, the FID value for the model without metadata has a significant change after $\text{epoch}=1500$ in ImageNet Church (bottom), which indicates that the model may break due to the instability of GANs. On the contrary, the curve of our full model is stable, which demonstrates that the implementation of metadata can stabilize the training process and prevent mode collapse.

Besides, as shown in Fig.~\ref{fig:plot}, compared with the model without the provision of metadata, our method can achieve a similar FID value but using much fewer epochs. For example, to reach the $\text{FID}=36.48$ in FFHQ (top), our full model only uses $50$ epochs, while the model without the metadata spends about $150$ epochs. This demonstrates that the adoption of metadata can speed up the training process, enabling the possibility to fast optimize the model in memory-limited devices.

\textbf{Effectiveness of the long-range module.}
To verify the effectiveness of the long-range module, we first conduct an ablation study, shown in Table~\ref{table:quantitative} and Fig.~\ref{fig:plot}. As we can see, without the implementation of the long-range module, the model has poor FID values in all three datasets.  
Also, as shown in Fig.~\ref{fig:plot}, the model without long-range module has significantly changeable FID values in FFHQ dataset (top red), and is hard to converge on ImageNet Church (bottom red), as the FID is still decreasing when $\text{epoch}=3500$, while our full model has achieved a better FID value at $\text{epoch}=1000$. This demonstrates that (1) our long-range module can effectively capture long-range dependencies to achieve a fast high-quality image generation, and (2) our proposed long-range module can work as a regularization to stabilize the training process and to prevent mode collapse.

Furthermore, to verify the performance of the long-range module, we conduct a comparison study that we use a residual block to replace our long-range module in the model (see Table~\ref{table:quantitative}), where the residual block has a similar number of parameters as the long-range module. 
We can easily observe that the model with the residual block has relatively worse FID values on three datasets. This comparison study demonstrates that the performance improvement achieved by using our long-range module is not simply because of an increase in model depth and capacity.

Besides, Table~\ref{table:quantitative} presents another comparison study, where the long-range module is replaced by self-attention. As we can see, compared with our full model, the model with self-attention has worse FID values on three datasets, and is even broken on FFHQ, because the FID value is 7 times larger than the value achieved by our method. This may indicate that the negative relationships preserved by our proposed long-range module can improve the performance of the model.

\section{Conclusion}
We have proposed novel lightweight long-range generative adversarial networks, which can efficiently generate realistic results without sacrificing image quality. More specifically, our model has a much smaller number of parameters and shorter inference time, but can still produce high-quality synthetic results. To achieve this, we have proposed a novel long-range module to capture long-range dependencies, which can also work as a regularization to prevent mode collapse. Besides, we have incorporated metadata into the image generation process to provide basic information about target images, which can stabilize the model and significantly speed up the training process. Extensive experimental results demonstrate the competitive performance of our method on three benchmark datasets, in terms of both visual fidelity and efficiency.

\newpage
\bibliography{iclr2021_conference}
\bibliographystyle{iclr2021_conference}

\newpage
\appendix

\section{Architecture}
As shown in Fig.~\ref{fig:archi}, the image encoder for metadata generation is a pretrained VGG-16~\citep{simonyan2014very} network. To extract more semantic information instead of fine contextual details, the deep neural network layer relu5\_3 of VGG-16 is adopted to derive the global visual representations, which contains more basic semantic information, such as the category and texture of objects, instead of fine-grained color, location, and shape information.

\subsection{Residual Block}
\begin{figure}[h!]
\centering
\begin{minipage}{0.199\textwidth}
\includegraphics[width=0.8\linewidth, height=1.42\linewidth]{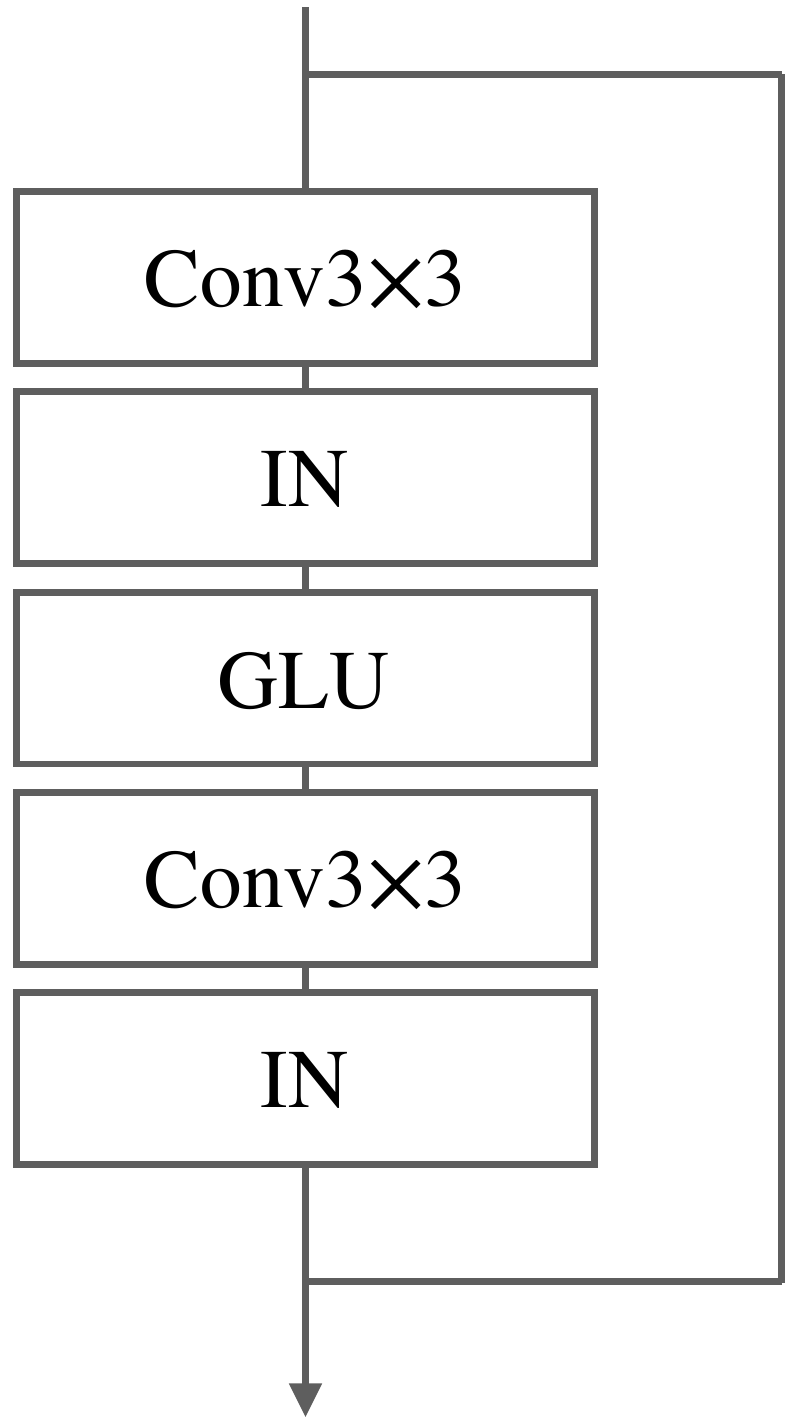}
\end{minipage}
\caption{Architecture of the residual block.}
\label{fig:archi_residual}
\end{figure}

Each residual block contains two convolutional layers, two instance normalizations (INs) \citep{ulyanov2016instance}, and one GLU \citep{dauphin2017language} non-linear function. The architecture of the residual block is shown in Fig.~\ref{fig:archi_residual}.

\subsection{Upsampling Block}
\begin{figure}[h!]
\centering
\begin{minipage}{0.199\textwidth}
\includegraphics[width=0.8\linewidth, height=1.42\linewidth]{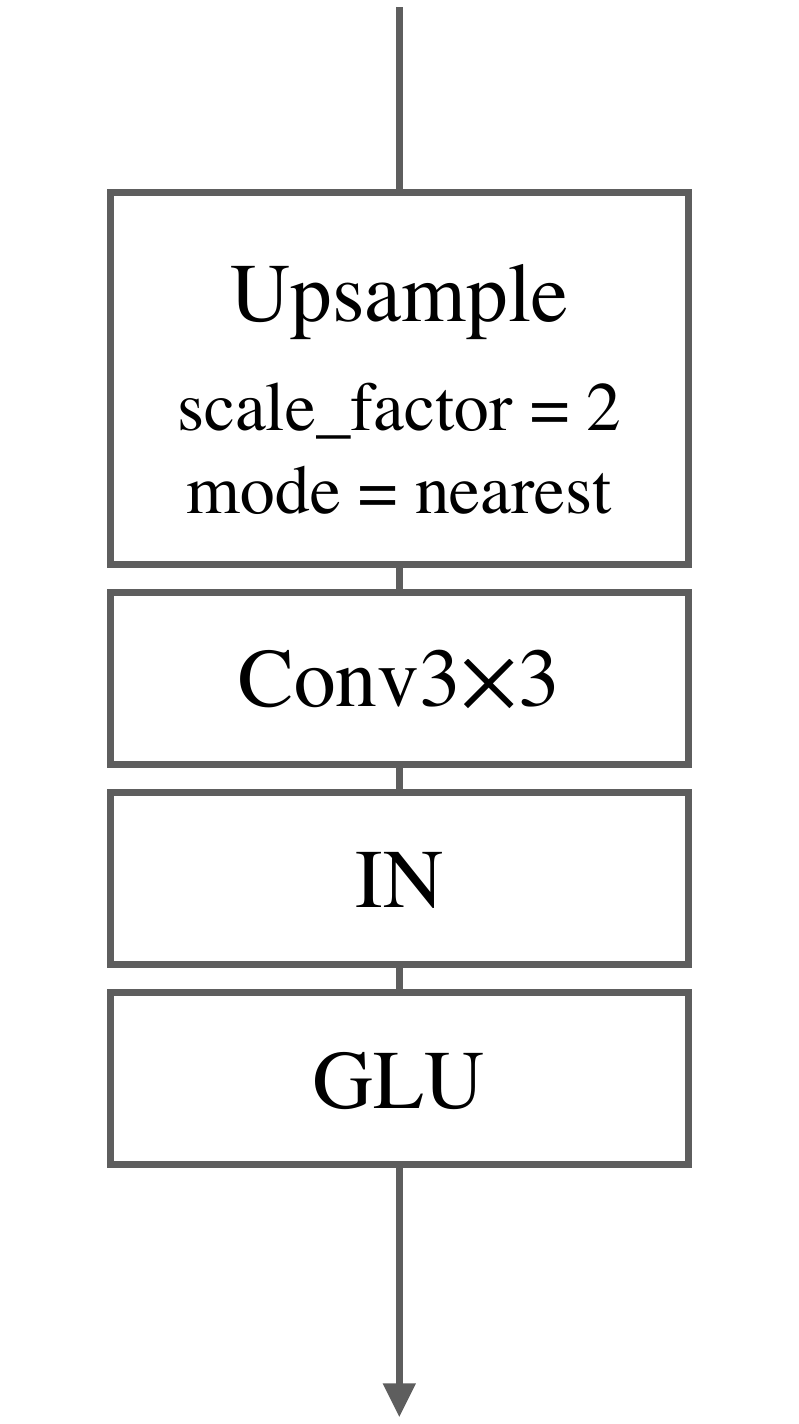}
\end{minipage}

\caption{Architecture of the upsampling block.}
\label{fig:archi_upsample}
\end{figure}

Each upsampling block contains one upsample function with nearest mode, one instance normalization (IN), one convolutional layer, and one GLU non-linear function. The architecture of the upsampling block is shown in Fig.~\ref{fig:archi_upsample}.

\section{Additional Results on FFHQ and CUB}
\begin{figure}[t]
\centering
\begin{minipage}{1\textwidth}
\includegraphics[width=1\linewidth, height=1.56\linewidth]{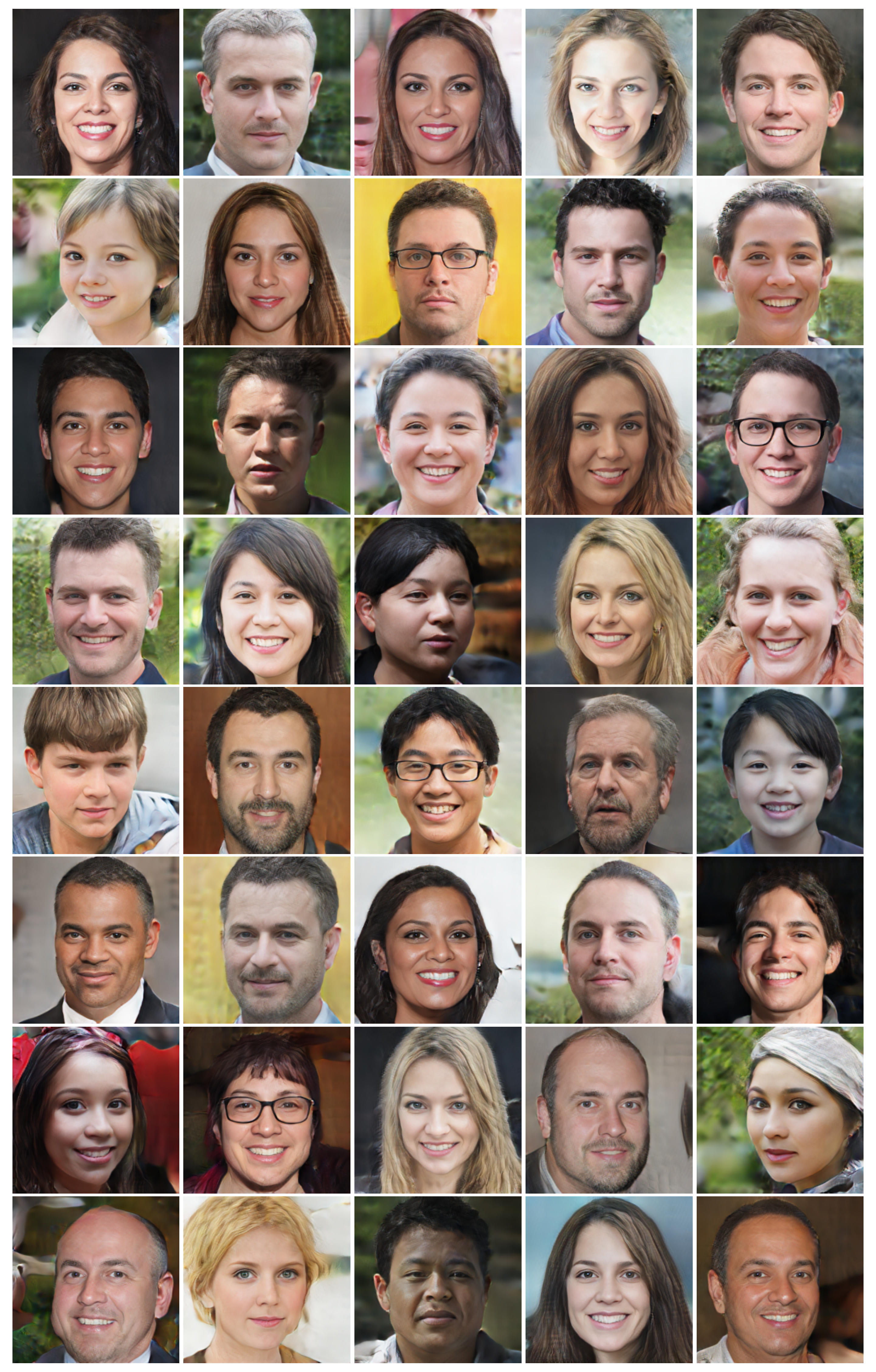}
\end{minipage}

\caption{Additional results generated by our method on the FFHQ dataset at $256 \times 256$.}
\label{fig:apex_face}
\end{figure}

\begin{figure}[t]
\centering
\begin{minipage}{1\textwidth}
\includegraphics[width=1\linewidth, height=1.56\linewidth]{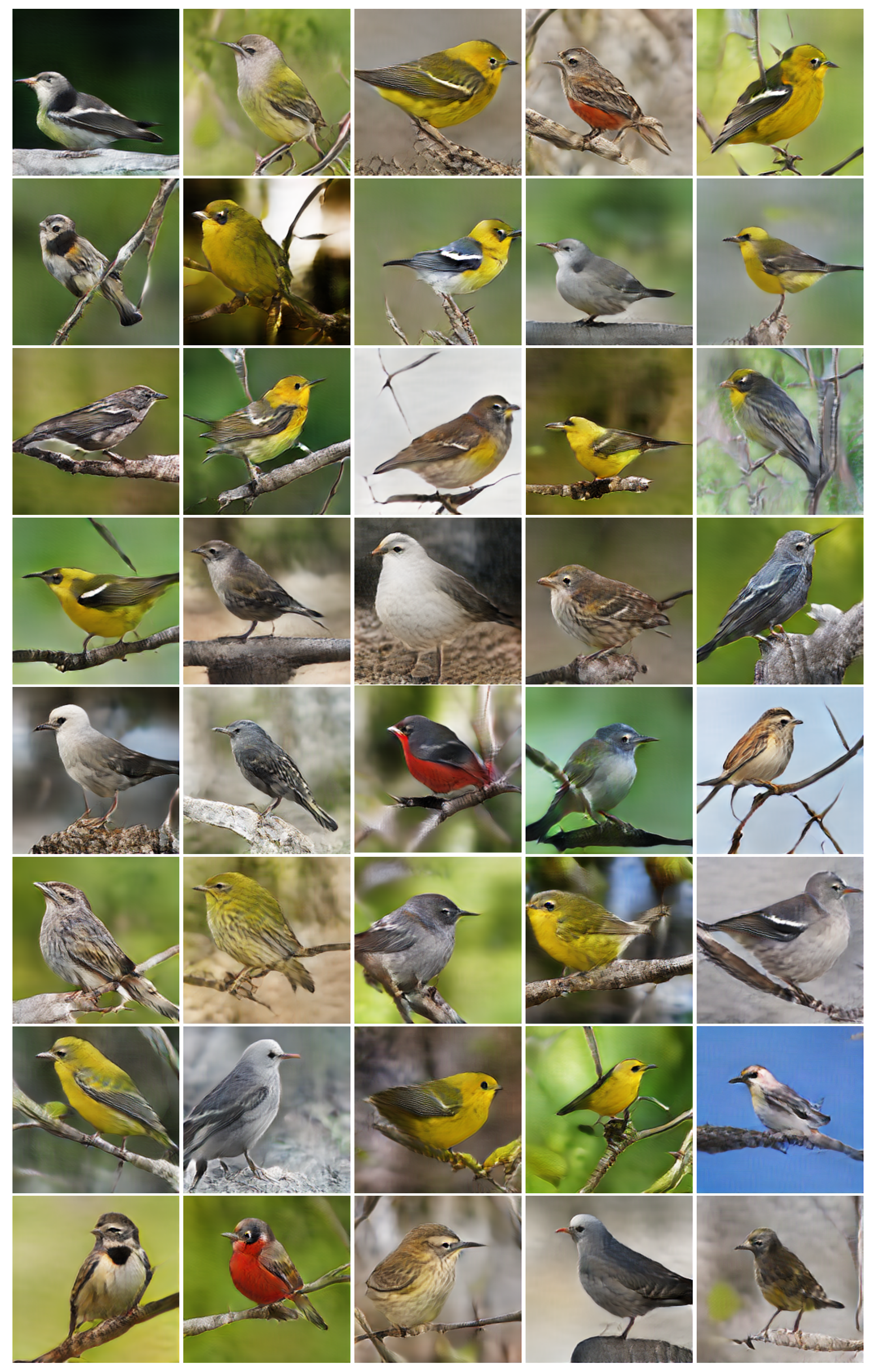}
\end{minipage}

\caption{Additional results generated by our method on the CUB bird dataset at $256 \times 256$.}
\label{fig:apex_bird}
\end{figure}

Figs.~\ref{fig:apex_face} and~\ref{fig:apex_bird} present additional results generated by our method on the FFHQ and CUB bird datasets, respectively.

\section{Additional ImageNet Results}
\begin{figure}[t]
\centering
\begin{minipage}{1\textwidth}
\includegraphics[width=1\linewidth, height=1.56\linewidth]{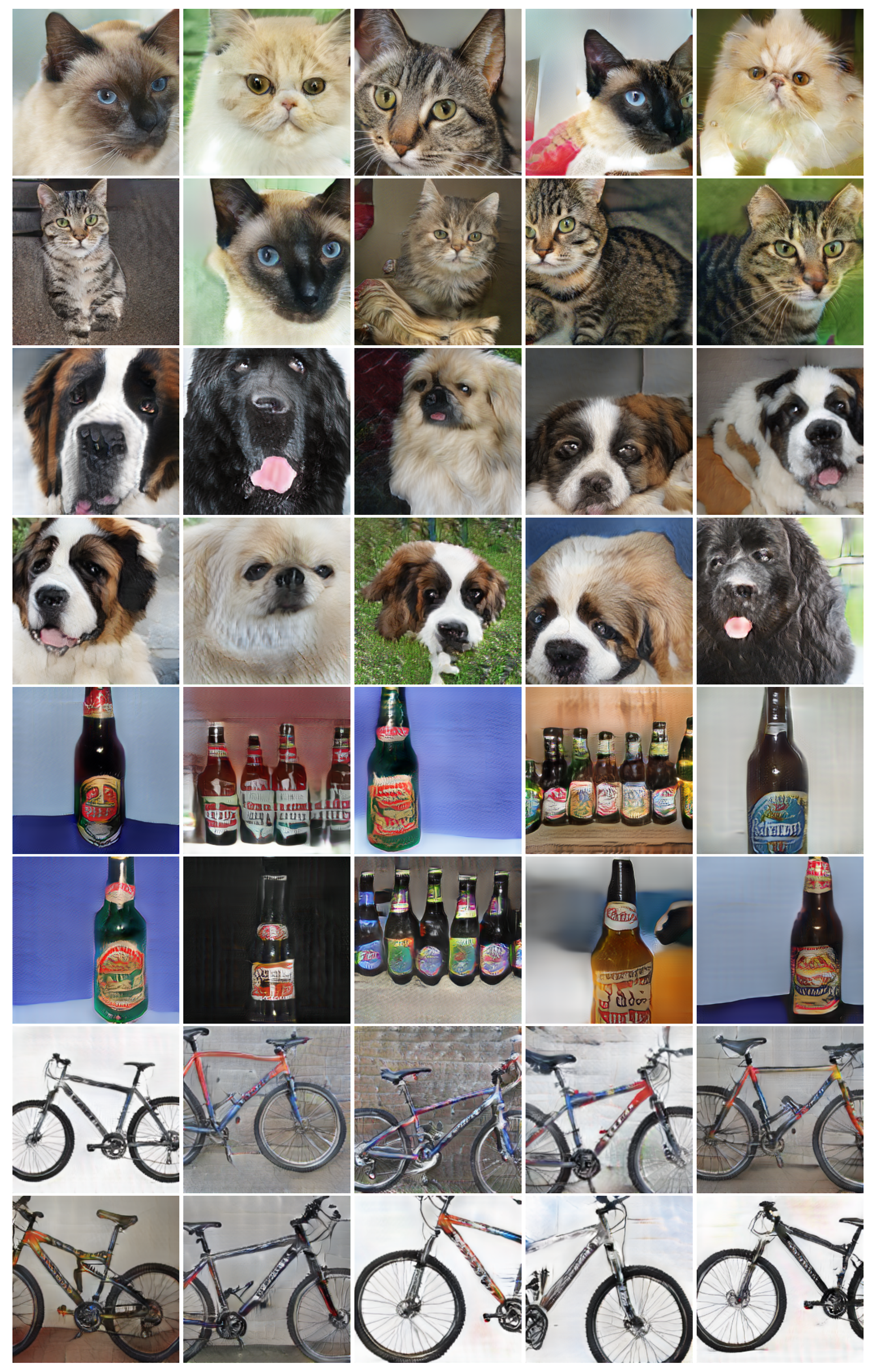}
\end{minipage}

\caption{Additional results generated by our method on the ImageNet dataset at $256 \times 256$. The categories are cat, dog, beer bottle, and bike.}
\label{fig:apex_imagenet1}
\end{figure}

\begin{figure}[t]
\centering
\begin{minipage}{1\textwidth}
\includegraphics[width=1\linewidth, height=1.56\linewidth]{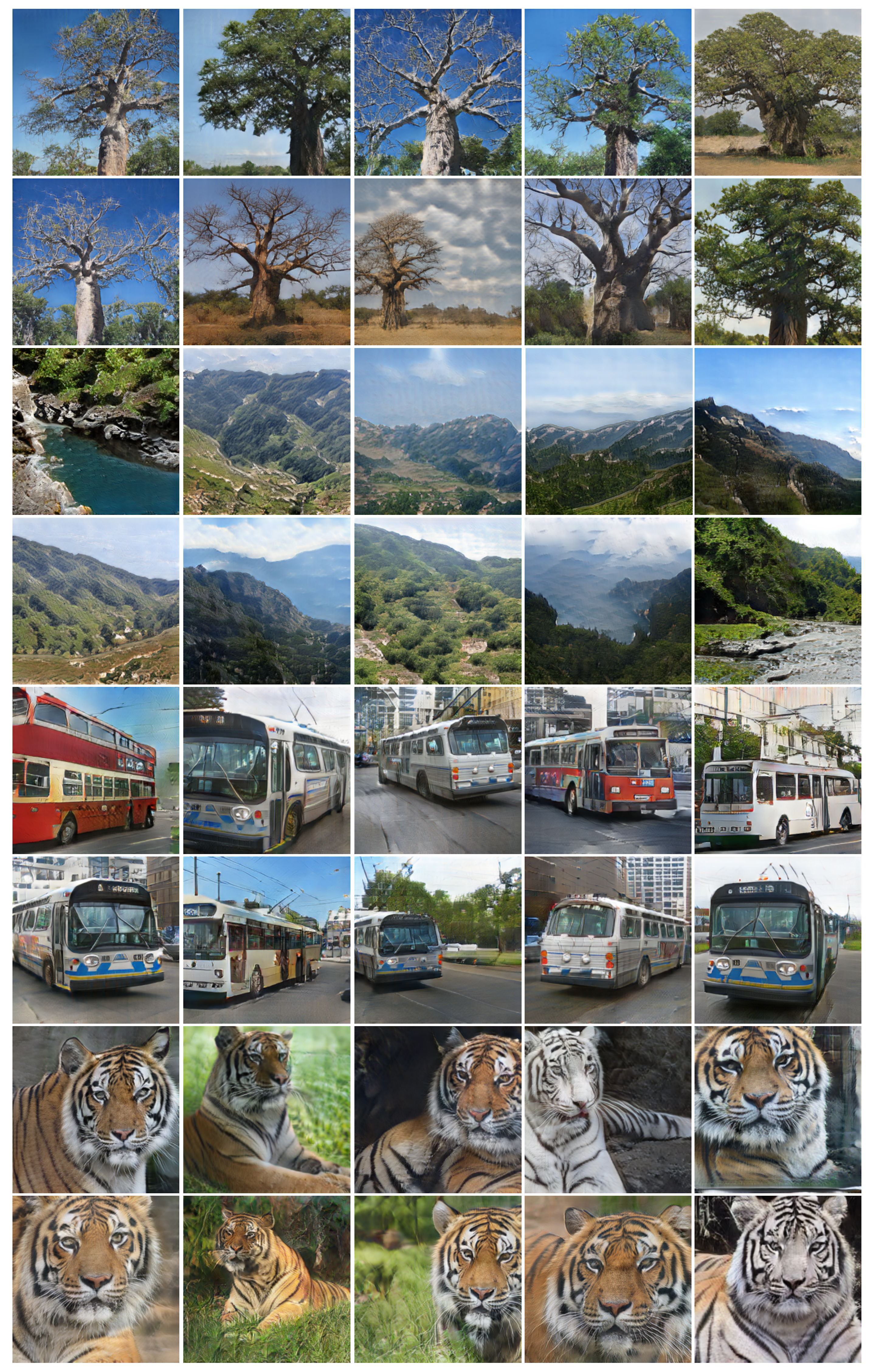}
\end{minipage}

\caption{Additional results generated by our method on the ImageNet dataset at $256 \times 256$. The categories are baobab, valley, bus, and tiger.}
\label{fig:apex_imagenet2}
\end{figure}

Figs.\ref{fig:apex_imagenet1} and~\ref{fig:apex_imagenet2} presents additional generated examples from the ImageNet dataset. A separate network is trained for each category using identical parameters.

\newpage

\newpage

\newpage

\newpage

\end{document}